# Homography Initialization and Dynamic Weighting Algorithm Based on a Downward-Looking Camera and IMU

Bo Dong, Yongkang Tao, Deng Peng, and Zhigang Fu

*Abstract*— In recent years, the technology in visual-inertial odometry (VIO) has matured considerably and has been widely used in many applications. However, we still encounter challenges when applying VIO to a micro air vehicle (MAV) equipped with a downward-looking camera. Specifically, VIO cannot compute the correct initialization results during take-off and the cumulative drift is large when the MAV is flying in the air. To overcome these problems, we propose a homography-based initialization method, which utilizes the fact that the features detected by the downward-looking camera during take-off are approximately on the same plane. Then we introduce the prior normal vector and motion field to make states more accurate. In addition, to deal with the cumulative drift, a strategy for dynamically weighting visual residuals is proposed. Finally, we evaluate our method on the collected real-world datasets. The results demonstrate that our system can be successfully initialized no matter how the MAV takes off and the positioning errors are also greatly improved.

## I. INTRODUCTION

VIO has been widely used in many fields, such as robots [1], unmanned aerial vehicles [2], and autonomous driving [3]. At present, many open-source VIO frameworks, which can be classified into tightly coupled [4]-[7] and loosely coupled [8, 9] approaches according to the method of information fusion, can satisfy the basic requirements of positioning systems. However, because of ever-changing scenes, these frameworks are normally incapable of coping with all robot localization problems. For example, if a MAV equipped with a downward-looking stereo camera uses the above-mentioned VIO systems when taking off from a parking apron, it may get wrong or inaccurate positioning results due to the failure of initialization.

Considering the high-speed motion and high-frequency vibration of the MAV during take-off, unstable initialization is unable to handle different scenes and as a result, the overall system may be prone to a malfunction. Therefore, it is necessary to design an effective and robust initialization method that can make the pose of the MAV converge to the optimal solution. In [10], authors propose a modified Martinelli-Kaiser solution [11] to deal with those features not seen by all cameras. To reduce time complexity, this method only uses a small set of detected features to optimize the gyroscope bias and gravity. Nevertheless, the method is not always effective, which is a hidden danger to VIO. Consequently, [12] introduces a three-step initialization strategy by the maximum-a-posteriori estimation. This strategy is based on visual-inertial bundle adjustment to further improve the accuracy of states. Moreover, it is also an

All authors are with the Centre for Flight Control and Navigation, Xpeng Aeroht, Guangzhou, China (e-mail: bodydonger@gmail.com; taoyk@aeroht.com; pengd@aeroht.com; 987324426@qq.com).

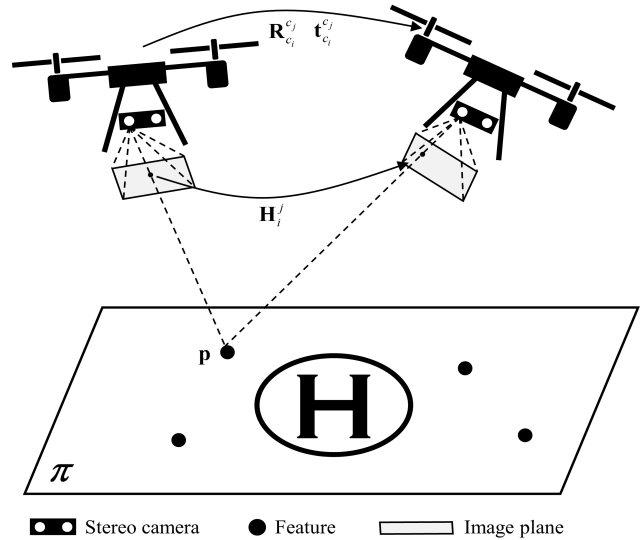

Fig. 1. Illustration of the take-off of a MAV from time $i$ to time $j$. All features are on the same plane $\pi$.

effective method for initialization to loosely couple the visual structure from motion and inertial measurement unit (IMU) pre-integration [13, 14]. The method computes states of VIO step by step, including scale factor, velocity, gravity, and gyroscope bias, which will be input to nonlinear optimization as initial values to improve the time consumption and accuracy. Also, given the influence of gravity, [15] refines its direction for better initialization results by using line features extracted by Line Segment Detector [16].

For the MAV equipped with a downward-looking stereo camera, because there is only a little intersection of the stereo camera's field of view when it is static on the ground, the VIO based on the above initialization cannot converge well. Given that the features detected by the downward-looking camera are almost on the same plane during take-off, a useful solution is to add homography to VIO [17]. In [18], authors utilize homography to assist a quadcopter in automatic landing. As the decomposition of a homography matrix can normally obtain four sets of solutions, authors design a method of selecting the correct one according to a prior normal vector. However, it should be noted that the above homography-based systems which suffer large cumulative drifts compute states with the help of other sensors like magnetometer, altimeter, and GPS, so they cannot directly be applied to a pure VIO system and achieve ideal performance.

Overall, despite a variety of methods, there are still no reasonable strategies to design a robust and accurate VIO system for a MAV equipped with a downward-looking stereo camera and IMU. Therefore, in this paper, we devise a novel method for solving the problem. Our main contributions are:

- A homography-based visual-inertial initialization method, which can compute states accurately.
- We introduce prior information and combine it with the motion of the MAV to select the correct rotation matrix and translation.
- We correlate the 2D normalized velocities of features with the 3D velocity of the body based on the homography constraint to further refine states.
- We propose a method of dynamically updating pixel deviation that can weight stereo visual residuals.

The rest of this paper is organized as follows. Section II introduces all notations of our system. The proposed methodology is described in Section III in detail. Section IV conducts experiments on our collected real-world datasets. Finally, conclusions are given in Section V.

## II. NOTATIONS

In this paper, the superscript and subscript $c$, $b$, and $w$ denote the camera, body, and world frame, respectively. $\mathbf{R}_A^B$, $\mathbf{t}_A^B$, and $\mathbf{T}_A^B \in SE(3)$ represent the rotation, translation, and pose of frame $A$ with respect to frame $B$. $\overline{\mathbf{t}}_A^B$ is the up-to-scale translation corresponding to $\mathbf{t}_A^B$. $\mathbf{v}_A^B$ is the velocity of the origin of frame $A$ viewed in frame $B$. $\mathbf{p}$ and $\overline{\mathbf{p}}$ are the 3D and homogeneous coordinate of a feature. $\mathbf{v}$ represents the velocity of a feature. $z$ denotes the depth of any feature. $f$ and $b_l$ are the focal length and baseline of a stereo camera. $(\hat{\bullet})$ denotes the estimated value of any variable. $\mathbf{H}$ is a homography matrix and can be partitioned into four submatrix blocks as

$$\mathbf{H} = \begin{bmatrix} \mathbf{h}_1^{2\times 2} & \mathbf{h}_2^{2\times 1} \\ \mathbf{h}_3^{1\times 2} & \mathbf{h}_4^{1\times 1} \end{bmatrix}, \quad (1)$$

where the superscript is the dimension of every submatrix.

## III. METHODOLOGY

### A. Overview

Our system follows the routine of [6] for detecting and tracking features. The take-off process of a MAV is shown in Fig. 1. No matter what its attitude is, those features detected by a stereo camera can be regarded as being on a plane $\pi$. As a result, we can compute the homography matrix $\mathbf{H}_i^j$ from time $i$ to time $j$ and then decompose it to obtain the rotation matrix $\mathbf{R}_{c_i}^{c_j}$ and up-to-scale translation $\overline{\mathbf{t}}_{c_i}^{c_j}$. It should be noted that the decomposition normally returns four different sets of solutions, only one of which is correct. Therefore, to restore the correct pose of the MAV, it is necessary to design a viable strategy to choose the correct one from different solutions. Considering that the normal vector of the plane where features are located corresponds exclusively to the true rotation matrix and translation, it is a feasible choice to introduce an initial vector as prior information to derive the correct solution. After that, as the homogeneous coordinate of a 3D feature loses its depth during projection from the camera to the normalized frame, the translation obtained by decomposing the homography matrix is up-to-scale. Because IMU can measure specific force, we solve the Perspective-n-Point (PnP) problem with its help to calculate the scale factor of the up-to-scale translation based on the least squares method. By now, we have obtained the accurate pose of the MAV, but there is another important state, velocity, which we have not optimized yet. Due to the low accuracy of consumer-grade IMU, there will be large positioning errors for initialization if solving velocity only relies on IMU. Fortunately, we find that homography can constrain not only a 3D feature's position but its velocity, so this is the premise to further refine velocity. Last but not least, to match stereo visual residuals with dynamic environments, we also propose a novel method for dynamically updating pixel deviation to improve positioning accuracy.

### B. Planar Homography Constraint

Assuming that a feature $\mathbf{p}$ is located on the plane $\pi$ and observed by camera $c_i$ at time $i$ and camera $c_j$ at time $j$, the picture is depicted in Fig. 1. The homogeneous coordinates of $\mathbf{p}$ in the normalized plane are $\overline{\mathbf{p}}_i$ and $\overline{\mathbf{p}}_j$. Then the homography constraint can be expressed as

$$\overline{\mathbf{p}}_j = \mathbf{H}_i^j \overline{\mathbf{p}}_i, \quad (2)$$

with

$$\mathbf{H}_i^j = \mathbf{R}_{c_i}^{c_j} + \frac{\overline{\mathbf{t}}_{c_i}^{c_j} \cdot \mathbf{n}_i^j}{d_i^j}, \quad (3)$$

where $\mathbf{n}_i^j$ is the normal vector of $\pi$. After obtaining the matched pairs of features between consecutive keyframes, $\mathbf{R}_{c_i}^{c_j}$ and $\overline{\mathbf{t}}_{c_i}^{c_j}$ can be computed according to (2) and (3). In our system, we use OpenCV [19] to perform this process. As mentioned above, $\overline{\mathbf{t}}_{c_i}^{c_j}$, proportional to the translation $\mathbf{t}_{c_i}^{c_j}$, is up-to-scale due to the loss of depth, that is, $\mathbf{t}_{c_i}^{c_j} = s\overline{\mathbf{t}}_{c_i}^{c_j}$.

### C. Normal-Based Selection Strategy

Decomposing $\mathbf{H}_i^j$ normally yields fours sets of $\mathbf{R}_{c_i}^{c_j}$ and $\overline{\mathbf{t}}_{c_i}^{c_j}$, only one of which is the correct solution. On the one hand, according to the condition that the depths of features are all positive, two sets of solutions can be excluded. On the other hand, distinguishing the remaining two sets must combine some man-made or environmental prior information. Assuming that the MAV remains stationary on $\pi$ at time $t_0$, since the image plane and $\pi$ are parallel to each other, the normal vector of $\pi$ can be represented as $\mathbf{n}_0 = [0, 0, 1]^T$ in the current camera frame. Considering that the downward-looking camera is very close to $\pi$ at this time, it is impossible to detect a sufficient number of features, so the initialization is not started for robustness. In contrast, we initialize VIO when the MAV reaches a pre-set height (normally less than 3m). Before that, we only depend on the IMU kinematic model to derive the trajectory of the MAV (as shown in Fig. 2). During the IMU-only process, the normal vectors between time $k$-1 and time $k$ are

$$\mathbf{n}_k = \mathbf{R}_{c_{k-1}}^{c_k} \mathbf{n}_{k-1}. \tag{4}$$

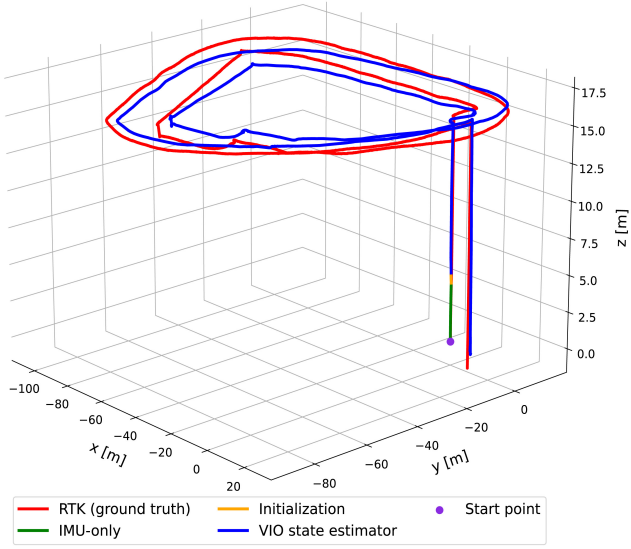

Fig. 2. A real-world experiment for our method. The MAV takes off at the start point where the downward-looking camera is unable to detect sufficient and high-quality features due to a little intersection of the stereo camera's field of view. Hence, the discrete-time IMU state propagation model is performed at first (green trajectory), until the MAV reaches a pre-set height. Immediately thereafter, our method takes the next sets of consecutive keyframes (up to the size of a sliding window) to perform initialization (orange trajectory). After this, the VIO state estimator runs for a long time (blue trajectory).

When the MAV flies to the pre-set height at time $t_K$, our system starts to initialize. For the two remaining sets of solutions, $S_K^a = \{\mathbf{R}_K^a, \overline{\mathbf{t}}_K^a, \mathbf{n}_K^a\}$ and $S_K^b = \{\mathbf{R}_K^b, \overline{\mathbf{t}}_K^b, \mathbf{n}_K^b\}$, our selection strategy is

$$S_K^{true} = \begin{cases} S_K^a & \|\mathbf{n}_K - \mathbf{n}_K^a\| \le \|\mathbf{n}_K - \mathbf{n}_K^b\| \\ S_K^b & \|\mathbf{n}_K - \mathbf{n}_K^a\| > \|\mathbf{n}_K - \mathbf{n}_K^b\| \end{cases}. \tag{5}$$

According to (5), we can always spot the correct rotation matrix and translation during the initialization step. It should be noted that if the number of features is inadequate, we will not use the prior normal vector to perform initialization but only count on IMU to compute the pose of the MAV.

### D. Scale Recovery with PnP

As mentioned above, the translation chosen by (5) is up-to-scale, which cannot be directly used for nonlinear optimization, so we must find a way to recover the scale $s$ of the translation. We all know that IMU measurements can be used to calculate the pose of the MAV in the real world, which is a good inspiration for us to relate the normalized translation $\overline{\mathbf{t}}_{c_i}^{c_j}$ to the real-world translation $\mathbf{t}_{c_i}^{c_j}$. Specifically, we make the most of IMU measurements to assist the initializer in computing the scale $s$. To begin with, the stereo baseline is used to obtain the 3D coordinates of features in the camera frame. As a result, we get many 3D-2D pairs of features and then construct a PnP problem that can roughly compute the pose of the camera frame with respect to the world frame $\mathbf{T}_{c_j}^w$ which is consistent with the scale of the real world, and then we can use $\mathbf{T}_{c_j}^w$ to align the pose $\mathbf{T}_{c_j}^{c_i}$ decomposed by the homography matrix, that is,

$$\mathbf{T}_{c_j}^w = \mathbf{T}_{b_i}^w \mathbf{T}_c^b \mathbf{T}_{c_j}^{c_i}. \tag{6}$$

Since $s$ is only related to translation, we focus on the translation component of (6). It can be expressed as

$$\begin{aligned} \mathbf{t}_{c_j}^w &= s\mathbf{R}_{b_i}^w \mathbf{R}_c^b \overline{\mathbf{t}}_{c_j}^{c_i} + \mathbf{R}_{b_i}^w \mathbf{t}_c^b + \mathbf{t}_{b_i}^w \\ \Rightarrow s\overline{\mathbf{t}}_{c_j}^{c_i} &= (\mathbf{R}_{b_i}^w \mathbf{R}_c^b)^T (\mathbf{t}_{c_j}^w - \mathbf{t}_{b_i}^w) - (\mathbf{R}_c^b)^T \mathbf{t}_c^b \\ \Rightarrow s\overline{\mathbf{t}}_{c_j}^{c_i} &= \hat{\mathbf{t}}_{c_j}^{c_i}. \end{aligned} \tag{7}$$

Due to the influence of the nonlinear model and biases in measurements, (7) does not hold strictly. Therefore, we use the least squares to solve $s$, which can be modeled as

$$\min_s \left\| s\overline{\mathbf{t}}_{c_j}^{c_i} - \hat{\mathbf{t}}_{c_j}^{c_i} \right\|. \tag{8}$$

After that, the rotation matrix and translation in the world frame can be recovered immediately.

### E. Velocity Update Based on Motion Field

Because the frequency of IMU is higher than that of camera, IMU measurements need to be integrated many times between every two consecutive keyframes. In this case, once the initial velocity of every keyframe has an error, then this error will be passed to the next keyframe and further increased, which may make the initializer break down. To overcome this problem, we consider optimizing velocity during the initialization stage. We rewrite (2) as [20]

$$\overline{\mathbf{p}}_j = \frac{\mathbf{h}_1 \overline{\mathbf{p}}_i + \mathbf{h}_2}{\mathbf{h}_3 \overline{\mathbf{p}}_i + \mathbf{h}_4}. \tag{9}$$

The derivatives of both sides of (9) are

$$\overline{\mathbf{v}}_j = \frac{(\mathbf{h}_3 \overline{\mathbf{p}}_i + \mathbf{h}_4)\mathbf{h}_1 - (\mathbf{h}_1 \overline{\mathbf{p}}_i + \mathbf{h}_2)\mathbf{h}_3}{(\mathbf{h}_3 \overline{\mathbf{p}}_i + \mathbf{h}_4)^2} \overline{\mathbf{v}}_i, \tag{10}$$

where $\overline{\mathbf{v}}_i$ and $\overline{\mathbf{v}}_j$ represent the normalized velocities of $\overline{\mathbf{p}}_i$ and $\overline{\mathbf{p}}_j$. Considering a feature $\mathbf{p}_c = [x, y, z]^T$ in the camera frame, its normalized coordinate is $\overline{\mathbf{p}} = [x/z, y/z, 1]^T$. Therefore, its normalized velocity can be represented as

$$\overline{\mathbf{v}} = \begin{bmatrix} \frac{1}{z} & 0 & -\frac{x}{z^2} \\ 0 & \frac{1}{z} & -\frac{y}{z^2} \end{bmatrix} \mathbf{v}_c, \tag{11}$$

where $\mathbf{v}_c$ is the velocity of $\mathbf{p}_c$. Since the MAV mainly performs vertical ascending motion during take-off, the rotational part of the camera can be approximately ignored. In this case, $\mathbf{v}_c$ can be obtained by

$$\mathbf{v}_c = -\mathbf{R}_b^c \mathbf{R}_w^b \mathbf{v}_c^w. \tag{12}$$

In our system, the camera and the IMU are rigidly connected, so we compute $\mathbf{v}_c^w$ by

$$\mathbf{v}_c^w = \mathbf{v}_b^w + \boldsymbol{\omega}_b \times \mathbf{t}_b^c, \quad (13)$$

where $\boldsymbol{\omega}_b$ is the angular velocity of the IMU.

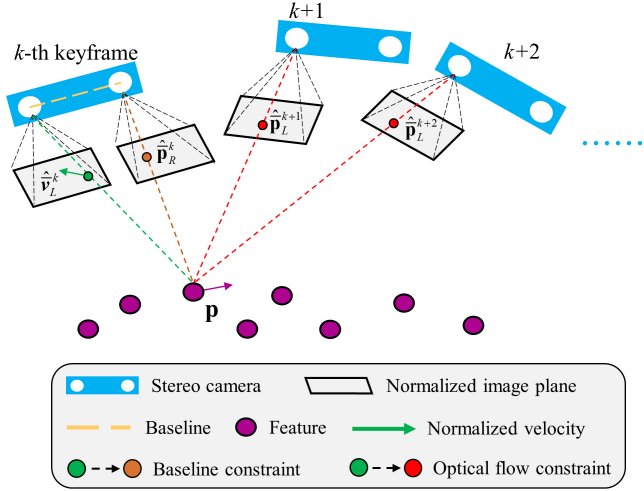

Fig. 3. An illustration of calculating the pixel deviation of a feature. The $k$-th keyframe is the one where the feature is observed for the first time.

According to (10), (11), (12), and (13), we can relate the normalized velocities to the 3D motion field and then utilize the Gauss-Newton method for optimizing $\mathbf{v}_b^w$, that is,

$$\min_{\mathbf{v}_b^w} \sum_j \left\| \overline{\mathbf{v}}_j - \hat{\mathbf{v}}_j \right\|. \quad (14)$$

Therefore, in the initialization stage, we depend on the homography constraint not only to derive the pose but to refine the velocity, which can make nonlinear optimization avoid getting stuck in the local optimum.

### F. Dynamically Weighting Stereo Visual Residuals

The weight of the visual residual is a negative correlation with the pixel deviation of any feature. In many open-source VIO frameworks, for example, VINS Mono [13] and VINS Fusion [21], the pixel deviation is fixed at 1.5 pixels, which means that all visual residuals have the same weight during nonlinear optimization. In general, feature detection and tracking threads are sensitive to different environments and devices, so it is not always reasonable to set the pixel deviation to a constant, which may impair the performance of VIO. We thus propose a novel method for dynamically updating the pixel deviation.

As shown in Fig. 3, for a feature $\mathbf{p}$, it can be observed by some consecutive keyframes, and the $k$-th keyframe is the one where $\mathbf{p}$ is observed for the first time. In this case, the stereo visual residuals include two parts:

1) Reprojection error from the left camera to right camera at time $k$.
2) Reprojection error from the left camera at time $k$ to the right camera at time $k+1$ or $k+2$.

For the first type of residual, we can obtain the pixel deviation of $\mathbf{p}$ with the help of the stereo baseline $b_l$. Assuming that its pixel coordinate is $(u_L^k, v_L^k)$ at time $k$, then its depth can be calculated by

$$z^k = \frac{f \cdot b_l}{u_L^k - u_R^k}. \quad (15)$$

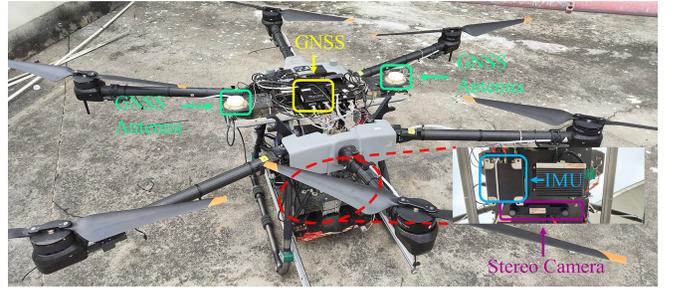

Fig. 4. The hexacopter equipped with an onboard NVIDIA AGX Xavier is used. The PPS signal from the GNSS receiver which can provide RTK solutions is used to align the camera time with global time.

Transforming $\overline{\mathbf{p}}_L^k$ to the right camera frame, we can get

$$\hat{\mathbf{p}}_R^k = z^k \cdot \overline{\mathbf{p}}_L^k + [-b_l,\ 0,\ 0]^T. \quad (16)$$

As a result, the pixel deviation corresponding to the first type of residual can be represented as

$$\sigma_{vis}^{(1)} = f \cdot \left\| \frac{\hat{\mathbf{p}}_R^k}{\hat{\mathbf{p}}_R^k(z)} - \overline{\mathbf{p}}_R^k \right\|. \quad (17)$$

For the second type of residual, we have to devise another method for computing the pixel deviation, which is because in this case, there is no rigid connection between cameras at different times. Considering that the optical flow field is an approximation of the 3D motion field, we can derive the pixel deviation via the motion field of the MAV. As shown in Fig. 3, assuming that there is a uniform motion between the two keyframes, if we know the normalized velocity of $\mathbf{p}$ at time $k$, then we can obtain its normalized coordinate and pixel deviation at time $k+1$. Specifically, the velocity $\mathbf{v}_c$ in the camera frame can be formulated as [22]

$$\mathbf{v}_c^k = -\mathbf{v}_{c_k}^{c_{k+1}} - \boldsymbol{\omega}_{c_k} \times \mathbf{p}_c^k, \quad (18)$$

where $\boldsymbol{\omega}_{c_k}$ is the angular velocity of the camera at time $k$. Therefore, given (11) and (18), the estimated normalized coordinate of $\mathbf{p}$ at time $k+1$ can be modeled as

$$\hat{\overline{\mathbf{p}}}_L^{k+1} = \overline{\mathbf{p}}_L^k + \hat{\overline{\mathbf{v}}}_L^k \cdot \Delta t_k. \quad (19)$$

Thus the pixel deviation is formulated as

$$\sigma_{vis}^{(2)} = f \cdot \left\| \hat{\overline{\mathbf{p}}}_L^{k+1} - \overline{\mathbf{p}}_L^{k+1} \right\|. \quad (20)$$

Based on (17) and (20), we can dynamically change the weight of the stereo visual residual to achieve the purpose of improving the accuracy and robustness of our system.

## IV. REAL-WORLD EXPERIMENTS

As illustrated in Fig. 4, the hexacopter is equipped with some sensors, including a downward-looking stereo camera, an IMU, and a GNSS receiver which can provide RTK solutions. Since the RTK solution has centimeter-level accuracy, we compare it with our results as the ground truth. The camera timestamp is synchronous with the global time by the PPS signal. The frequency and resolution of the stereo camera are 20Hz and 1280×800 respectively. The frequency of IMU is 200Hz. The algorithm is implemented in C++ and runs on Ubuntu 20.04 and ROS noetic.

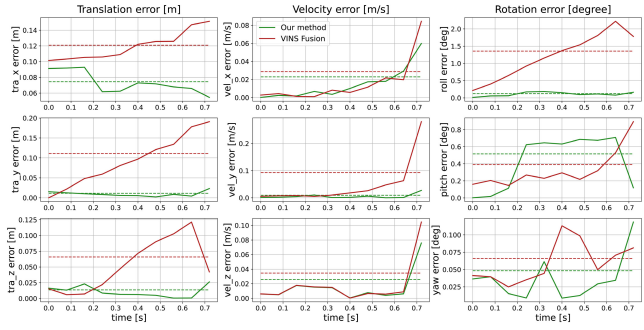

Fig. 5. The error in our method and VINS Fusion within a sliding window during initialization for states, including translation, velocity, and rotation. The dashed line is the root mean squared error (RMSE) for every component.

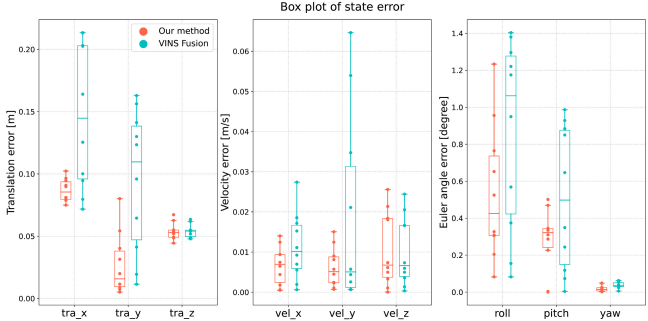

Fig. 6. Box plots of translation, velocity, and Euler angle error in our method and VINS Fusion during initialization.

*A. Initialization Results*

Firstly, we analyze the positioning performance within a sliding window during initialization on our collected dataset. The results are plotted in Fig. 5, which includes translation, velocity, and Euler angle errors. We compare our system with an open-source system, VINS Fusion, which is also a stereo VIO system but performs initialization with pure PnP. For the three components of translation, we see that our errors are less than 0.1m in all three directions, and even less than 0.05m in $y$ and $z$ directions. Compared with VINS Fusion, our proposed method reduces the RMSE by 38.4%, 89.9%, and 79.3% in $x$, $y$, and $z$ axes. To analyze the rotation, we transform the rotation matrix into Euler angles in the north-east-down frame. Our method outperforms VINS Fusion in terms of roll and yaw directions. The roll angle, in particular, fluctuates slightly around the ground truth and its error is very close to zero during the whole initialization stage.

To further analyze the performance of the two methods, we visualize the box plot of the error on another collected data, as shown in Fig. 6. We can see that our method has smaller errors in all three states, and the distributions of errors are more concentrated, which makes it possible for our system to perform nonlinear optimization more smoothly.

*B. Different Take-off Attitude*

Some challenging scenes may degrade the performance of visual algorithms, such as weak textures or dark scenes.

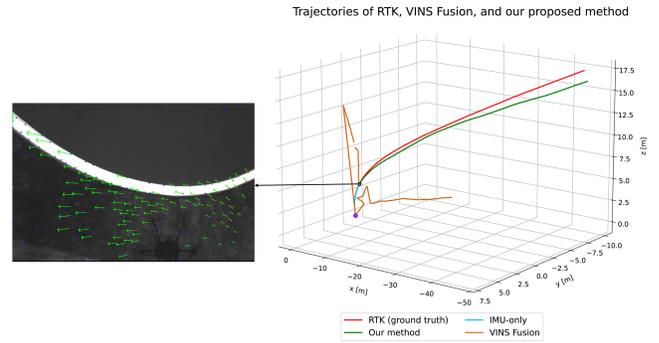

(a). Oblique Take-off

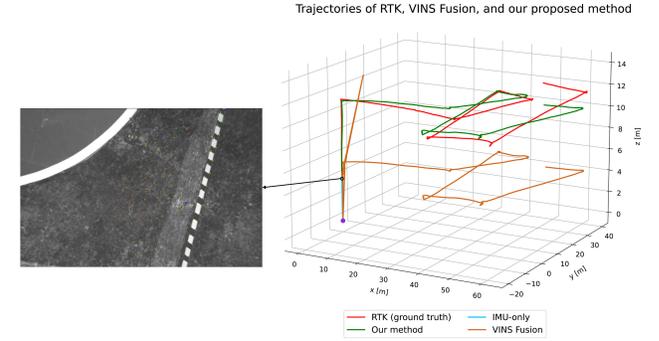

(b). Vertical Take-off

Fig. 7. Effects of different take-off attitudes on initialization. (a) Oblique Take-off: The stereo camera is close to the ground and the distribution of detected features is uneven. (b) Vertical Take-off: The scene texture is not obvious, so the number of detected features is insufficient.

Therefore, we conduct experiments under two rigorous take-off attitudes, that is, oblique and vertical take-off, to evaluate the performance of our method, as shown in Fig. 7. When the MAV takes off obliquely, VIO starts initialization at a close distance from the ground. At this time, some scenes lack textures, which leads to the uneven distribution of features. In Fig. 7(a), due to the smooth surface of the parking apron on which the MAV is located, VIO cannot detect the features in this area, so the detected features are all concentrated in the lower half of this keyframe, which is why VINS Fusion fails. Specifically, the uneven distribution of features makes the pose calculated by PnP inaccurate, so there is a large error in gyroscope bias calibration, which causes the failure of initialization. In contrast, our proposed method can perform initialization successfully under such a degraded scene. Since we add the homography constraint to our system, it can limit the feasible region of the states of our system to a great extent during the initialization step, so that all states are more likely to be near the ground truth. Even in the case of the uneven distribution of features, the homography constraint can still characterize planar structures and constrain poses, which ensures that our method can perform well in any scene. Another example of vertical take-off also demonstrates the validity of our method as shown in Fig. 7(b). We see that, in this scene, the detected features are all distributed in the middle area of the keyframe, with few on both sides. Our method, however, successfully performs initialization for the same reason above.

*C. Long-Term Positioning Error*

Although our method has a good performance in the initialization step, we are more concerned about whether the system can still perform well when entering nonlinear optimization. We thus conduct experiments on the collected dataset which lasts about 290s to verify the long-term positioning accuracy. As shown in Fig. 8, VIO completes initialization about 20s after startup and then runs nonlinear optimization. Although the accuracy of our method is only slightly better than that of VINS Fusion in the initialization step, our advantage becomes more obvious over time after the system performs the back-end thread due to the addition of dynamic pixel deviation. The error in $z$ axis, in particular, has always a smaller cumulative drift than that of VINS Fusion which increases over time. This shows the superiority of our method when it comes to long-term positioning.

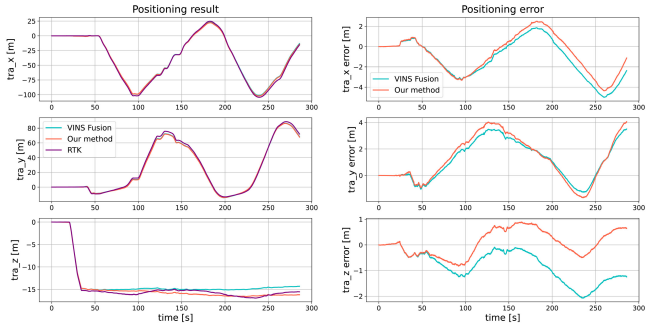

Fig. 8. The positioning result and error of VINS Fusion and our proposed method after initialization.

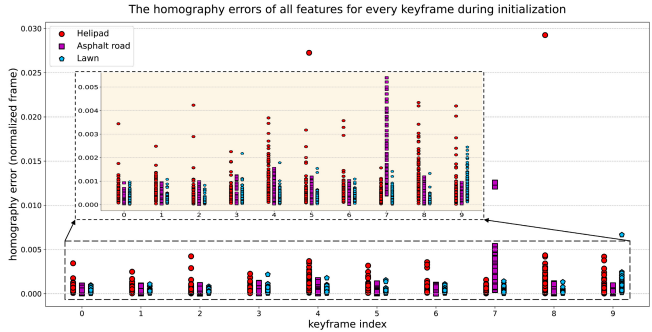

Fig. 9. Performance of our homography-based initialization method in three different scenes (helipad, asphalt road, and lawn).

*D. Initialization Performance in Different Take-off Scenes*

As mentioned above, the homography constraint requires that all features are on the same plane, which, however, is difficult to be satisfied in the real world. Therefore, we evaluate the performance of our proposed initialization in three different take-off scenes (helipad, asphalt road, and lawn). We choose the homography reprojection error as the indicator, that is, for any feature,

$$ind_i^j = \left\| \bar{\mathbf{p}}_j - \mathbf{H}_i^j \bar{\mathbf{p}}_i \right\|. \tag{21}$$

As shown in Fig. 9, we can see that during initialization, any feature error is less than 0.03. Furthermore, almost all feature errors are less than 0.006, and only a few features (actually 5) have relatively large errors. Of the three scenes, the helipad is the most similar to a smooth plane, followed by the asphalt road. In contrast, the lawn has the least characteristic of the smooth plane, but our method can still successfully perform initialization and achieve accurate positioning results, which proves the robustness of our method in the real world.

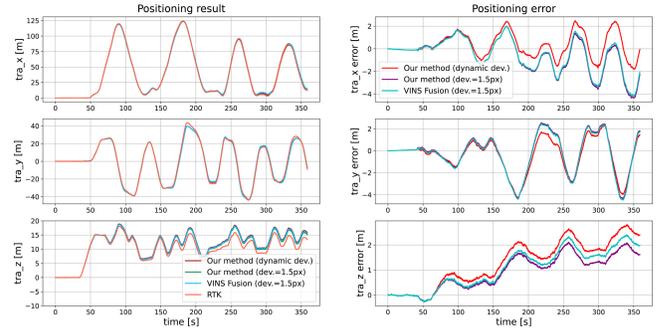

Fig. 10. The three components of the trajectory and error in our method and VINS Fusion. The fixed pixel deviation is set to 1.5 pixels. When we dynamically update the pixel deviation, the cumulative errors in $x$ and $y$ axes are greatly improved.

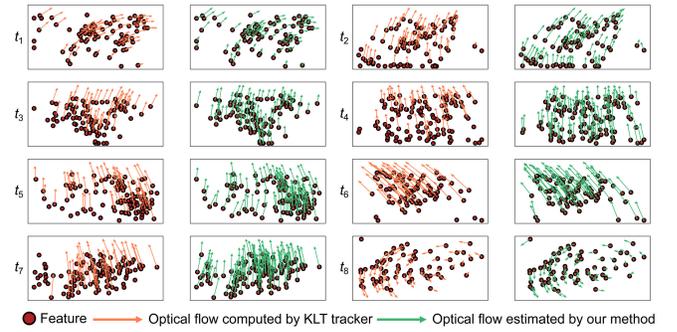

Fig. 11. A comparison between the optical flow computed by the KLT tracker and estimated by our method at eight different times.

*E. Performance of Dynamically Weighting Visual Residuals*

In this experiment, we test how the pixel deviation affects the positioning accuracy. As shown in Fig. 10, when the pixel deviation is set to a fixed value of 1.5 pixels, the cumulative drift in $x$ axis increases over time, which is because feature detection and tracking are sensitive to illumination, so those features cannot have the same pixel deviation. By contrast, after we perform the dynamic weighting strategy, our results are greatly improved. The maximum error in $x$ axis has reduced from 4m to 2m and the fluctuation range of the error curve has also become smaller. We also compare the optical flow estimated by our dynamic algorithm with that computed by the Kanade-Lucas-Tomasi (KLT) tracker as depicted in Fig. 11. We can see that the two are consistent whether the MAV is flying in translational or rotational motion.

## V. CONCLUSION

In this paper, we present a novel initialization method, which can deal with the problem that the traditional VIO is prone to failure when a MAV equipped with a downward-looking camera takes off. Specifically, since the features detected by the downward-looking camera are approximately on the same plane during the take-off of the MAV, we use homography to compute the inter-frame pose. Then we introduce the prior normal vector and motion field to make the pose converge to the ground truth. Additionally, we find that when the MAV flies in the air for a long time, VIO suffers large cumulative drifts. To handle this, a dynamic weighting algorithm is proposed, which can effectively alleviate the cumulative drifts. Through experimental results on the collected real-world datasets, we highlight that our proposed method is effective, reliable, and robust.